\documentclass[runningheads]{llncs}
\usepackage[T1]{fontenc}
\usepackage{graphicx}
\usepackage{amsmath}
\usepackage{amsfonts}
\usepackage{xcolor}

\begin{document}
\title{Wasserstein Barycenter Gaussian Process based Bayesian Optimization}
\titlerunning{WBGP-BO}
%
\author{Antonio Candelieri\inst{1}\orcidID{0000-0003-1431-576X} \and
Andrea Ponti\inst{1}\orcidID{0000-0003-4187-4209} \and
Francesco Archetti\inst{2}\orcidID{0000-0003-1131-3830}}
\authorrunning{A. Candelieri et al.}
%
\institute{Department of Economics Management and Statistics, University of Milano-Bicocca, 20126, Italy \and
Department of Computer Science Systems and Communication, University of Milano-Bicocca, 20126, Italy
\email{\{antonio.candelieri,francesco.archetti\}@unimib.it},\\
\email{a.ponti5@campus.unimib.it}}
\maketitle              
\begin{abstract}
Gaussian Process based Bayesian Optimization is a widely applied algorithm to learn and optimize under uncertainty, well-known for its sample efficiency. However, recently -- and more frequently -- research studies have empirically demonstrated that the Gaussian Process fitting procedure at its core could be its most relevant weakness. Fitting a Gaussian Process means tuning its kernel's hyperparameters to a set of observations, but the common Maximum Likelihood Estimation technique, usually appropriate for learning tasks, has shown different criticalities in Bayesian Optimization, making theoretical analysis of this algorithm an open challenge. Exploiting the analogy between Gaussian Processes and Gaussian Distributions, we present a new approach which uses a prefixed set of hyperparameters values to fit as many Gaussian Processes and then combines them into a unique model as a Wasserstein Barycenter of Gaussian Processes. We considered both "easy" test problems and others known to undermine the \textit{vanilla} Bayesian Optimization algorithm. The new method, namely Wasserstein Barycenter Gausssian Process based Bayesian Optimization (WBGP-BO), resulted promising and able to converge to the optimum, contrary to vanilla Bayesian Optimization, also on the most "tricky" test problems. 

\end{abstract}
\section{Rationale and motivation}

Bayesian Optimization (BO) is a sample-efficient model-based sequential method for the global optimization of black-box, multi-extremal, and expensive to evaluate objective functions \cite{archetti2019bayesian,frazier2018bayesian,garnett2023bayesian}. Its sample efficiency is the crucial factor of its successful application in many domains, like Automated Machine Learning \cite{candelieri2022fair,karras2023automl,sorokin2023sigopt}, robotic \cite{berkenkamp2023bayesian,zhang2023bayesian}, and  drug discovery \cite{colliandre2023bayesian}.

Without loss of generality, the reference problem considered in this paper is the following global minimization problem:
\begin{equation}
    \boldsymbol{x}^* \in \underset{\boldsymbol{x}\in\mathcal{X}\subset\mathbb{R}^h}{\arg\min}\; f(\boldsymbol{x})
    \label{eq:1}
\end{equation}
where $\mathcal{X}$ is the so-called \textit{search space}, usually assumed to be the $h$-dimensional unit hypercube $\mathcal{X}=[0,1]^h$, and with $f:\mathcal{X}\rightarrow \mathbb{R}$ some measurable objective to minimize, showing the properties previously mentioned.\\

A generic iteration of the \textit{vanilla} BO algorithm consists of (\textit{i}) generating an approximation of $f(\boldsymbol{x})$ depending on the observations collected so far -- denoted by $\mathcal{D}^{(n)}=(\mathbf{X},\mathbf{y})$, with $\mathbf{X}=\{\boldsymbol{x}^{(i)}\}_{i=1:n}$ and $\mathbf{y}=\{y^{(i)}\}_{i=1:n}$ -- and (\textit{ii}) choosing the next \textit{query} $\boldsymbol{x}^{(n+1)}$ balancing between exploration and exploitation. Specifically, $y^{(i)}=f(\boldsymbol{x}^{(i)})$ in the noise-free setting, otherwise $y^{(i)}=f(\boldsymbol{x}^{(i)})+\varepsilon^{(i)}$ with $\varepsilon^{(i)}$ usually assumed as a zero-mean Guassian noise.\\

Gaussian Process (GP) regression \cite{gramacy2020surrogates,williams2006gaussian} is the most common choice for approximating $f(\boldsymbol{x})$ depending on $\mathcal{D}^{(n)}$, leading to the so-called GP-based BO.

Fitting a GP model to a set of observations means tuning the GP kernel's hyperparameters, usually via Maximum Likelihood Estimation (MLE). Another option is Maximum A Posteriori (MAP), that is just a \textit{penalized} MLE: the only difference is that the likelihood is weighted with respect to some prior.

However, as recently and increasingly reported in the literature, this procedure is affected by different issues. First, MLE in GP fitting is known to be ill-posed \cite{karvonen2023maximum}. The most common and practical workaround consists into \textit{injecting} additional noise (aka \textit{nugget effect}) to the available observations, even if the target problem is known to be noise-free, at the cost of a degradation in the GP model's precision. Second, MLE is consistent only if the observations fill the space uniformly, which is definitely the opposite setting of BO, where observations should concentrate close to the global optimizer \cite{hvarfner2024self}. Finally, and more importantly, theoretical convergence proofs of GP-based BO algorithms hold only in the \textit{standard setting} \cite{bogunovic2021misspecified}, in which one assumes a \textit{realizable}, also said \textit{well-specified} scenario, meaning that $f(\boldsymbol{x})$ is a member of a specific Reproducing Kernel Hilbert Space (RKHS) with bounded norm, known a-priori.

In simpler terms, convergence to the optimum is proved under the assumption that the GP kernel’s hyperparameters are known in advance, that is definitely an impractical assumption for real-life problems.

The most widely used GP-based BO algorithm is based on the GP's Confidence Bound \cite{srinivas2012information} and is widely quoted for being \textit{no regret}.\footnote{Denote with $R_N = \sum_{i=1}^N \left(f(x^{(i)})-f(x^*)\right)$ the \textit{regret} accumulated by an algorithm generating $N$ solutions. The algorithm has no-regret if $\lim_{N\rightarrow\infty} \frac{R_N}{N} \rightarrow 0$.} It is \textit{optimistic} in the face of uncertainty, with Lower Confidence Bound (GP-LCB) and Upper Confidence Bound (GP-UCB) separately used for minimization and maximization problems.
Like for all the other no regret BO algorithms, GP fitting via MLE leads the GP Confidence Bound to get stuck into local optima \cite{bull2011convergence}. Heuristic methods \cite{berkenkamp2019no,wabersich2016advancing,wang2014theoretical,wang2016bayesian} have been proposed over time, basically using the Lemma 4 of \cite{bull2011convergence}, stating that decreasing the \textit{length-scale} hyperparameter of the kernel leads to consider a larger RKHS, increasing exploration at the expense of a higher regret \cite{garnett2023bayesian}.

In other terms, all these methods aim at increasing the exploration with respect to the scheduling proposed in \cite{srinivas2012information} whenever the target scenario is miss-specified (i.e., not realizable). Miss-specification is so common in real-life settings to be considered the reason for the better performances provided by recent acquisition functions when compared against GP-LCB \cite{benjamins2022pi,berk2020randomised,de2021greed}.


Instead of using a predefined strategy for iteratively decreasing the length-scale, \cite{hvarfner2024self} has recently proposed SCoreBO (Self-Correcting Bayesian Optimization) which extends Statistical distance-based Active Learning (SAL) to perform BO and active learning, simultaneously. As a result, SCoreBO provides better estimates of GP kernel's hyperparameters compared to vanilla BO algorithms, outperforming them on a set of benchmarks. On the other hand, SCoreBO’s efficiency is definitely contingent to the type of kernel adopted: when the number of hyperparameters values increases over time instead of converge, this means that $f(\boldsymbol{x})$ does not belong to the set of functions representable through the chosen kernel. Moreover, the self-correcting mechanism entails a large use of computational resources.
More recently, \cite{ziomek2024beyond} proposed HE-GP-UCB (Hyperparameter Elimination GP-UCB) which eliminates hyperparameters values that are highly implausible depending on the observations. The next query is an optimistic choice with respect to the not yet eliminated hyperparameters values. Another approach has been presented\footnote{Presented last year at LION 18.} in \cite{candelieri2024mle}, namely MLE-free GP-BO, which is not only a more effective and efficient BO algorithm, but it also represents a novel framework in which GP fitting, ill-conditioning, and exploration-exploitation trade-off are simultaneously addressed by tuning the GP kernel’s hyperparameters without using MLE.\\

In this paper we present a novel approach which uses a prefixed set of values for the GP kernel's hyperparameters to fit as many GPs and then combines them into a unique (GP) model obtained as a Wasserstein Barycenter of GPs, where Wasserstein refers to a distance metric between probability distributions. Thus, contrary to the other state-of-the-art approaches, we do not modify or select the GP kernel's hyperparameters values over iteration but just combine them proving that such a combination is statistically grounded. We first summarize the methodological background about GP based BO and the Wasserstein distance. Then, we detail our algorithm and present some preliminary experimental results. Finally, we summarize our relevant conclusions.

\section{Methodological background}

\subsection{Gaussian Process and Lower Confidence Bound}
A GP can be intuitively thought of as a generalization of a Gaussian Distribution (GD) over functions (instead of scalar values). A sample from a GP is a collection of random variables such that any finite restriction of their values has a joint GD. Similarly to a GD, a GP is completely defined by its mean and covariance, which are functions over $\boldsymbol{x}$. Fitting a GP regression model means conditioning these two functions to a set of observations -- i.e.,  $\mathcal{D}^{(n)}$ in BO -- leading to the following two equations denoting the GP's predictive mean and variance.
\begin{equation}
    \begin{split}
        \mu(\boldsymbol{x}) & = \mu_0(\boldsymbol{x}) + \mathbf{k}(\boldsymbol{x},\mathbf{X}) \left[\mathbf{K}+\sigma_\varepsilon^2 \mathbf{I}\right]^{-1} (\mathbf{y}-\boldsymbol{\mu}_0(\mathbf{X}))\\        
        \sigma^2(\boldsymbol{x}) & = k(\boldsymbol{x},\boldsymbol{x}) - \mathbf{k}(\boldsymbol{x},\mathbf{X}) \left[\mathbf{K}+\sigma_\varepsilon^2\mathbf{I}\right]^{-1} \mathbf{k}(\mathbf{X},\boldsymbol{x})
    \end{split}    
\end{equation}

with $\mu_0(\boldsymbol{x})$ the prior mean (usually assumed to be zero) and $\boldsymbol{\mu}_0(\mathbf{X})$ a shorthand for $\left(\mu_0(\boldsymbol{x}^{(1)}),...,\mu_0(\boldsymbol{x}^{(n)})\right)$. Then, $k(\boldsymbol{\cdot,\cdot})$ is a \textit{kernel function} which is used as the GP's \textit{covariance function}, thus $\mathbf{k}(\boldsymbol{x},\mathbf{X})$ is a vector whose $i$th component is $k(\boldsymbol{x},\boldsymbol{x}^{(i)})$ and $\mathbf{K}$ is an $n \times n$ matrix with entries $\mathbf{K}_{ij}=k(\boldsymbol{x}^{(i)},\boldsymbol{x}^{(j)})$. Finally, $\sigma_\varepsilon^2$ is the variance of a zero-mean Gaussian noise which is considered in the case that $f(\boldsymbol{x})$ is noisy or \textit{injected} to avoid ill-conditioning in the inversion of $\mathbf{K}$.\\

Given the current $\mu(\boldsymbol{x})$ and $\sigma(\boldsymbol{x})$ (i.e., the squared root of $\sigma^2(\boldsymbol{x})$), the next query is identified by optimizing an \textit{acquisition function} which balances between exploration and exploitation. According to our reference problem (\ref{eq:1}), we consider the GP-LCB acquisition function, that is:
\begin{equation}
    \boldsymbol{x}^{(n+1)} \in \underset{\boldsymbol{x} \in \mathcal{X}}{ \arg\min }\; \mu(\boldsymbol{x}) - \xi \sigma(\boldsymbol{x})
\end{equation}
with $\xi>0$ regulating the trade-off between exploitation and exploration (i.e., a large value of $\xi$ lead to a high \textit{uncertainty bonus}, moving towards exploration).

\subsection{Wasserstein distance}
When there is the need to measure how much two probability distributions differ, statistical divergences are typically used, such as Kullback-Liebler, Jensen-Shannon, and $\chi 2$, depending on the type of distributions (i.e., discrete or continuous). However, it is well-known that statistical divergences are not \textit{distances} because they do not satisfy all the required properties. In Statistics and Machine Learning the so-called Wasserstein distance has increasingly gained importance during last years: it allows to compare two probability distributions -- even of different types -- satisfying all the properties of a distance, under certain assumptions. For a comprehensive discussion on the Wasserstein distance, the reader can refer to \cite{ollivier2014optimal,peyre2019computational,santambrogio2015optimal}; in this paper we just summarize the background related to the scope of the paper. As far as the relevance of Wasserstein distance in the ML community is concerned, a relevant example is given in \cite{arjovsky2017wasserstein} and in two recent surveys on the topic \cite{montesuma2024recent,khamis2024scalable}.\\

Denote by $\Omega$ an arbitrary space and $d$ a metric on that space. Let $\mathcal{P}(\Omega)$ be the set of Borel probability measures on $\Omega$ and $\alpha$ and $\beta$ two probability measures in $\mathcal{P}(\Omega)$, then their $p$-Wasserstein distance is defined as follows \cite{cuturi2014fast}:
\begin{equation}
    \mathcal{W}_p^p(\alpha,\beta) = \underset{\pi \in \Pi(\alpha,\beta)}{\inf} \int_{\Omega \times \Omega} d^p(\omega,\omega') d\pi(\omega,\omega')
    \label{eq:wst}
\end{equation}
where $\Pi(\alpha,\beta)$ is the set of probability measures on $\Omega \times \Omega$ having marginals $\alpha$ and $\beta$, and $p\geq1$.
Setting $p=2$ and $d$ the Euclidean distance leads to the so-called 2-Wasserstein distance with equation (\ref{eq:wst}) that can be rewritten as
\begin{equation}
    \mathcal{W}^2_2(\alpha,\beta) = \underset{T\natural\alpha = \beta}{\min}\int_\Omega \|T(\omega)-\omega\|^2_2 \; d\alpha(\omega)
    \label{eq:W2}
\end{equation}
which is intended as an optimal transport problem \cite{peyre2019computational}: the \textit{source} probability mass $\alpha$ must be moved to match the target probability mass $\beta$ at the minimum cost. Specifically, the transportation cost is computed with respect to a \textit{ground metric}, that is the Euclidean distance in (\ref{eq:W2}). The symbol $\natural$ denotes the so-called \textit{push-forward operator}, meaning that following its transport according to the transport plan $T$, $\alpha$ must match $\beta$. The minimizer of (\ref{eq:W2}), typically denoted with $T^*$, is named optimal transport plan.\\

Specific results are available in the case that $\alpha$ and $\beta$ are GDs, and they will be particularly useful in this paper, given the analogy between GDs and GPs, also discussed in \cite{mallasto2017learning}.
If $\alpha = \mathcal{N}(\mathbf{m}_\alpha,\boldsymbol{\Sigma}_\alpha)$ and $\beta = \mathcal{N}(\mathbf{m}_\beta,\boldsymbol{\Sigma}_\beta)$, then the computation of their 2-Wasserstein distance simply becomes
\begin{equation}
    \mathcal{W}^2_2(\alpha,\beta) = \|\mathbf{m}_\alpha - \mathbf{m}_\beta\|^2 + \mathcal{B}(\boldsymbol{\Sigma}_\alpha,\boldsymbol{\Sigma}_\beta)^2
    \label{eq:W2_gauss}
\end{equation}
where $\mathcal{B}$ is the Bures metric \cite{bures1969extension} between positive definite matrices \cite{forrester2016relating}, that is:
\begin{equation}
    \mathcal{B}(\boldsymbol{\Sigma}_\alpha,\boldsymbol{\Sigma}_\beta)^2 = \text{trace} \left(\boldsymbol{\Sigma}_\alpha + \boldsymbol{\Sigma}_\beta - 2 (\boldsymbol{\Sigma}_\alpha^{1/2} \boldsymbol{\Sigma}_\beta \boldsymbol{\Sigma}_\alpha^{1/2})^{1/2}\right)
\end{equation}

Thus, in the case of centered GDs (i.e., $\mathbf{m}_\alpha=\mathbf{m}_\beta=\boldsymbol{0}$), the 2-Wasserstein distance resembles to the Bures metric. Moreover,  if $\boldsymbol{\Sigma}_\alpha$ and $\boldsymbol{\Sigma}_\beta$ are diagonal, the Bures metric is the Hellinger distance -- which is considered along with Wasserstein in \cite{hvarfner2024self} for the SAL component of SCoreBO. Finally, in the commutative case, that is $\Sigma_\alpha\Sigma_\beta=\Sigma_\beta\Sigma_\alpha$, the Bures metric is equal to the Frobenius norm $\|\Sigma_\alpha^{1/2}-\Sigma_\beta^{1/2}\|^2_{Frobenius}$.\\

For the scope of this paper, we are specifically interested in the setting consisting of univariate GDs, which allows to further simplify equation (\ref{eq:W2_gauss}) into:
\begin{equation}
    \mathcal{W}_2^2(\alpha,\beta)=(m_\alpha-m_\beta)^2 + (\sigma_\alpha - \sigma_\beta)^2
        \label{eq:W2univar}
\end{equation}
Indeed, in the case of univariate GDs, $\mathcal{W}_2^2$ is the squared Euclidean distance between two GDs represented as points into a bi-dimensional space having the mean and the standard deviation as dimensions.

\subsection{Wasserstein Barycenter}
Given a suitable distance between probability measures, a first possible application consists into computing the mean of a set of them, that is their Wasserstein barycenter.
This is a variational problem involving all Wasserstein distances from one to many measures, with the aim to identify the probability measure that minimizes the sum of its Wasserstein distances to each element in that set. 
 
Formally, given a set of probability measures $\{\alpha_i\}_{i=1}^N$ in $\mathbb{P} \subset P(\Omega)$ with associated weights $\{\lambda_i\}_{i=1}^N$ with $\sum_{i=1}^N$,  their Wasserstein barycenter $\bar\alpha$ is:
\begin{equation}
    \bar\alpha \in \underset{\alpha \in \mathbb{P} \subset P(\Omega)}{ \arg \min } \sum_{i=1}^N \lambda_i\mathcal{W}_2^2(\alpha,\alpha_i)
\end{equation}
where $\{\lambda_i\}_{i=1}^N$ are called \textit{barycentric coordinates}. The most common situation consists into considering the probability measures as equally weighted, meaning that $\lambda_i=1/N, \forall i=1,...,N$ -- and leading to
\begin{equation}
    \bar\alpha \in \underset{\alpha \in \mathbb{P} \subset P(\Omega)}{ \arg \min } \frac{1}{N} \sum_{i=1}^N \mathcal{W}_2^2(\alpha,\alpha_i)
\end{equation}

According to (\ref{eq:W2univar}), if the probability distributions are univariate GDs, that is $\alpha_i = \mathcal{N}(m_i, \sigma_i)$, then $\bar\alpha=\mathcal{N}(\bar{m},\bar{\sigma})$ with
\begin{equation}
    \bar{m}, \bar{\sigma} \in \underset{m, \sigma}{ \arg \min } \frac{1}{N} \sum_{i=1}^N \Big[(m_i - m)^2 + (\sigma_i - \sigma)^2 \Big]
    \label{eq:11}
\end{equation}
whose solution is simply given by $\bar{m}=\frac{1}{N}\sum_{i=1}^N m_i$ and $\bar{\sigma}=\frac{1}{N}\sum_{i=1}^N \sigma_i$.\\

\begin{figure}[h!]
    \centering
    \includegraphics[width=0.7\linewidth]{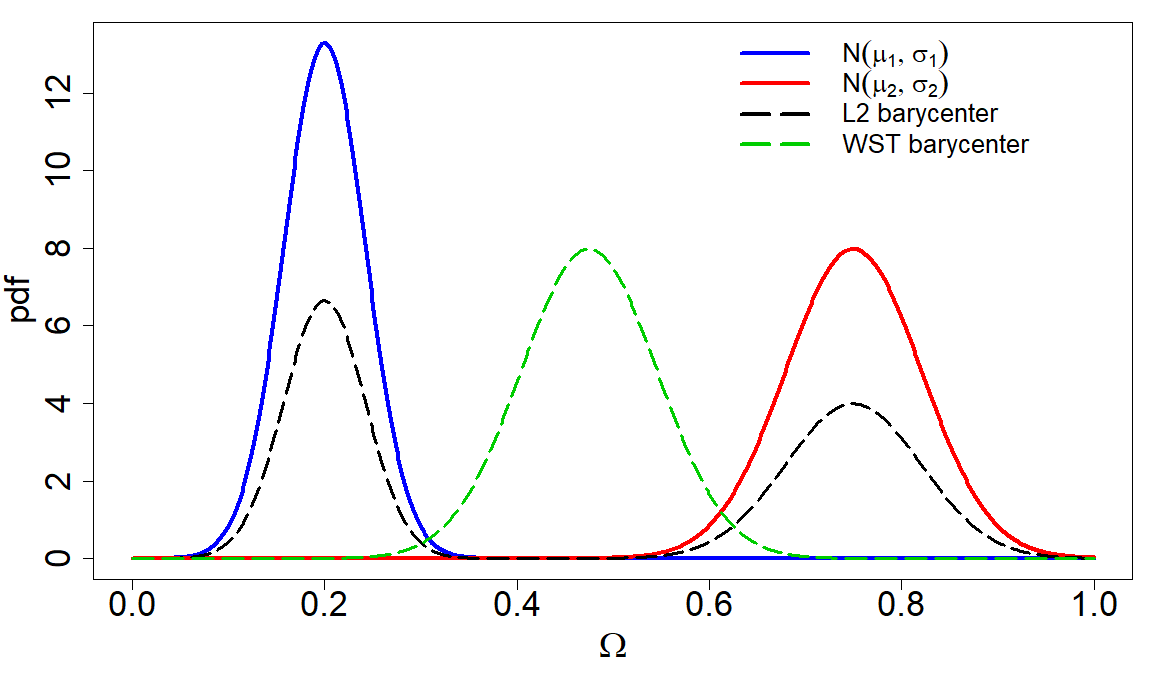}
    \caption{Shape preservation of the Wasserstein barycenter against L2-norm barycenter.}
    \label{fig:WST_bary}
\end{figure}

Figure \ref{fig:WST_bary} provides a practical and simple example showing the difference between the Wasserstein barycenter of two normal distributions and the L2-norm barycenter obtained by considering as distance the L2-norm between the probability density functions aligned on the support. The \textit{shape preservation} guarateed by the Wasserstein barycenter is clearly visible: it is normal instead of bimodal.
\vspace{-0.25cm}

\section{The WBGP-BO algorithm}

The algorithm proposed in this paper, namely WBGP-BO (Wasserstein Barycenter Gaussian Process based Bayesian Optimization) stems from the simple consideration that given to a GP fitted on the available observations $\mathcal{D}^{(n)}$ the expected value of $f(\boldsymbol{x})$, at any location $\boldsymbol{x} \in \mathcal{X}$, is just:
\begin{equation}
    f(\boldsymbol{x}) \sim \mathcal{N}(\mu(\boldsymbol{x}),\sigma(\boldsymbol{x}))
    \label{eq:12}
\end{equation}

Assume to have $N$ different GPs, each one with its own kernel's type or hyperparameters, then we have $N$ different $\mu_i(\boldsymbol{x})$ and $\sigma_i(\boldsymbol{x})$ to use for (\ref{eq:12}). Since all of them are reasonable, it could be advantageous to sample from their Wasserstein barycenter, that is:
\begin{equation}
    f(\boldsymbol{x}) \sim \mathcal{N}(\bar{\mu}(\boldsymbol{x}),\bar{\sigma}(\boldsymbol{x}))
\end{equation}
with
$\bar{\mu}(\boldsymbol{x})=\frac{1}{N}\sum_{i=1}^N \mu_i(\boldsymbol{x})$ and $\bar{\sigma}(\boldsymbol{x})=\frac{1}{N}\sum_{i=1}^N\sigma_i(\boldsymbol{x})$, according to (\ref{eq:11}).\\

\begin{figure}[h!]
    \centering
    \includegraphics[width=0.8\linewidth]{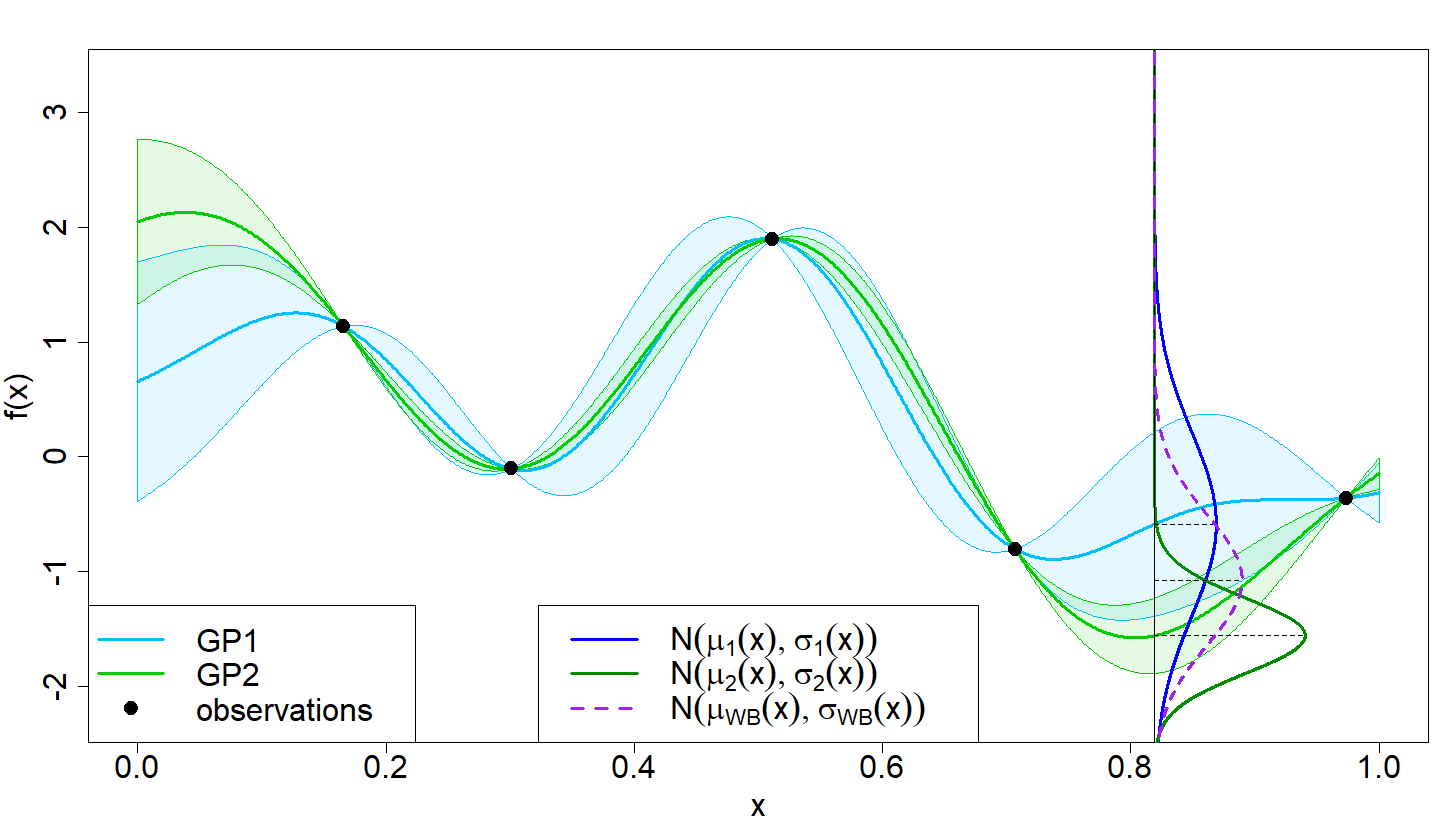}
    \caption{Two different GPs fitted to the same set of five observations. On the vertical line: the two Gaussian distributions (blue and green) for posterior about $f(\boldsymbol{x})$, at that specific location, and their associated Wasserstein barycenter (purple, dashed).}
    \label{fig:WBGP_x}
\end{figure}

Figure \ref{fig:WBGP_x} shows two different GPs (same kernel but different length scale) fitted to the same set of five observations. Although \textit{locally} similar, they provide different possible estimates of $f(\boldsymbol{x})$ over the entire search space. On the vertical line, the two posterior Gaussian probability measures for $f(\boldsymbol{x})$ at that specific location are depicted, along with their Wasserstein barycenter.
\vspace{-0.25cm}

\begin{figure}[h!]
    \centering
    \includegraphics[width=0.8\linewidth]{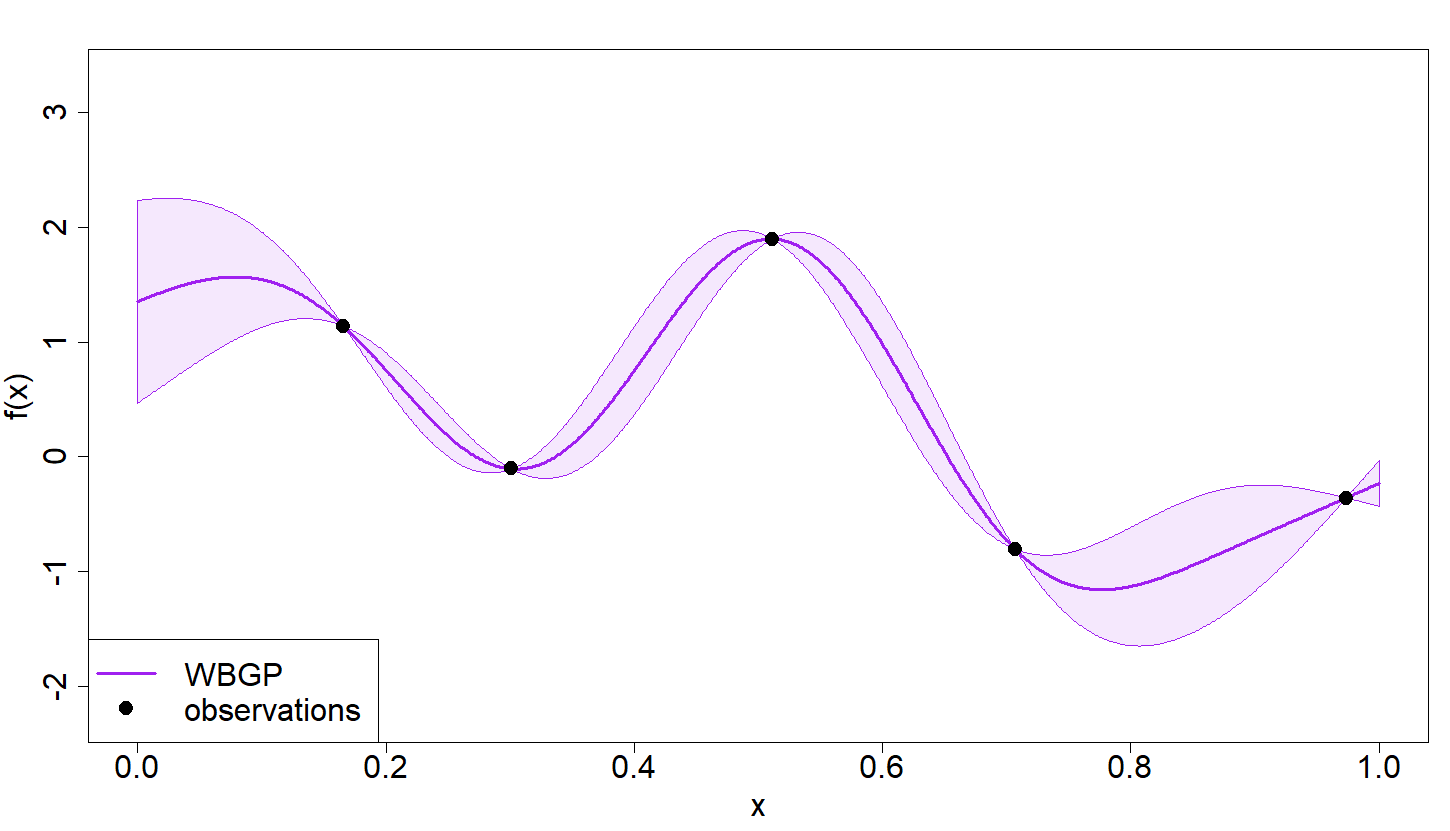}
    \caption{The Wasserstein Barycenter Gaussian Process (WBGP) obtained by computing the Wasserstein barycenter of the two GPs in Figure \ref{fig:WBGP_x}, $\forall\;\boldsymbol{x} \in \mathcal{X}$.}
    \label{fig:WBGP}
\end{figure}

When the computation of the Wasserstein barycenter is performed over the entire search space, the Wasserstein Barycenter Gaussian Process (WBGP) can be depicted, as reported in Figure \ref{fig:WBGP}. It encompasses structural properties from both the two original GPs, even if it is different from both of them.

According to what discussed so far, and considering the GP-LCB acquisition function, we can postulate the following Theorem.\\

\noindent
\textbf{Theorem}. \textit{Given $N$ different GPs fitted to the current set of observations $\mathcal{D}^{(n)}$, the average of their LCBs, at a given location $\boldsymbol{x}\in \mathcal{X}$, is equal to the LCB of the Wasserstein barycenter of those GPs at  $\boldsymbol{x}$, if $\xi_i=\xi\;\forall i=1,...,N$.}
\begin{equation}
    \underbrace{\frac{1}{N} \sum_{i=1}^N \Big[\mu_i(\boldsymbol{x}) - \xi \sigma_i(\boldsymbol{x})\Big]}_\text{average of LCBs at $\boldsymbol{x}$} = \frac{1}{N} \sum_{i=1}^N \mu_i(\boldsymbol{x}) - \frac{1}{N} \sum_{i=1}^N\xi \sigma_i(\boldsymbol{x}) = \underbrace{\bar{\mu}(\boldsymbol{x}) - \xi \bar{\sigma}(\boldsymbol{x})}_\text{WBGP's LCB at $\boldsymbol{x}$}
\end{equation}

\noindent
\textbf{Proposition}.\textit{Obviously, the previous theorem holds for the upper confidence bound, too: it is sufficient to replace $-\xi$ with $+\xi$}.\\

Another possible, but quite practical, interpretation for the WBGP is to think of it as an ensemble of GPs, with the final prediction computed as a (equally) weighted average of the individual predictions but the final uncertainty obtained as a (equally) weighted average of the individual uncertainties. This specific ensemble coincides with a WBGP-BO and is a statistically grounded combination of GPs.

\section{Experiments and results}
In this paper, experiments are just limited to a set of one-dimensional test problems, which allowed us to better investigate the behaviour of the WBGP-BO algorithm with respect to vanilla BO.

\begin{figure}[h!]
    \centering
    \includegraphics[width=1\linewidth]{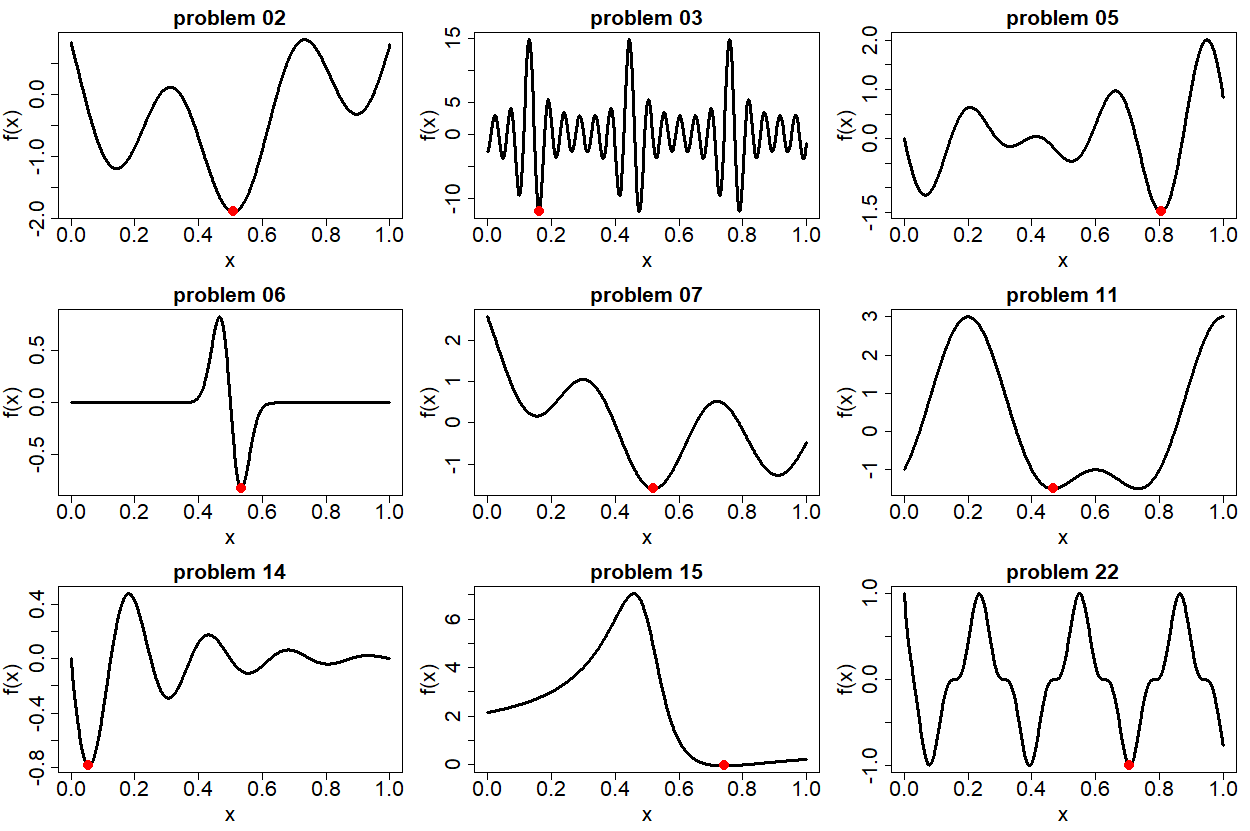}
    \caption{Graphical representation of the test problems reported in Table \ref{tab:1} (the original search space of each function has been rescaled in $[0,1]$).}
    \label{fig:test_functions}
\end{figure}

We have selected nine test problems associated to different settings where GP-BO (i.e., vanilla BO with GP fitting through MLE, and GP-LCB as acquisition function) can easily converge to the optimizer, in some cases, or irreversibly get stuck into a local minimum, in others. Table \ref{tab:1} summarizes the main characteristics of the test problems, while Figure \ref{fig:test_functions} provides a graphical representation of each one of them (with the original search spaces all rescaled to [0,1]). Intuitively, problems 03, 05, 06, 14, and 22, are the most "tricky" for GP-BO because the risk of missing the optimizer, due to a miss-specified GP, is significantly high.\\ 

For WBGP-BO we have used only one type of kernel, specifically the Squared Exponential (SE) kernel (aka Gaussian kernel), that is $k(x,x')=\sigma_f^2 e^{-\frac{(x-x')^2}{2\ell^2}}$. The set of $N$ different GPs is obtained by sampling $N$ values for the kernel's hyperparameters $\sigma_f^2$ and $\ell$. Two different settings have been considered, separately: $N=16$ and $N=32$, with values sampled from a large pool of 64 predefined possible $\sigma_f^2$-$\ell$ pairs (with $(\sigma_f^2, \ell) \in[0.01,0.5]^2$ according to a uniform $8\times 8$ grid). Also GP-BO uses a SE kernel, but MLE is used to tune its hyperparameters at each iteration.\\

Experiments are organized as follows: 5 initial observations sampled through Latin Hypercube Sampling (LHS) and 30 sequential iterations. To mitigate the impact of the random initialization, 30 different runs have been performed. For each run, GP-BO and WBGP-BO (with both $N=16$ and $N=32$) share the same randomly initialized set of observations.
\begin{table}[h!]
    \centering
    \resizebox{\textwidth}{!}{
    \begin{tabular}{|c|c|c|c|}
        \hline
        \textbf{Name} & $f(x)$ & $\mathcal{X}$ & \textbf{Solution} \\
        \hline \hline
        \;problem 02\; & $\sin(x) + \sin(\frac{10}{3}x)$ & [2.7, 7.5] &  $x^*=5.1457$; $f(x^*)=-1.8996$ \\
        problem 03 & $-\sum_{i=0}^5 \left[i\;\sin\left( (i+1) x + i\right)\right]$ & [-10, 10] & $x^*=-6.7746$; $f(x^*)=-12.0312$ \\
        problem 05 & $-(1.4-3x)\sin(18x)$ & [0, 1.2] & $x^*=0.9661$; $f(x)=-1.4891$\\
        problem 06 & $-[x+\sin(x)]e^{-x^2}$ & [-10, 10] & $x^*=0.6796$; $f(x^+)=-0.8242$\\ 
        \;problem 07\; & \;$\sin(x)+\sin(\frac{10}{3}x)+\log(x)-0.84x+3$\; & [2.7, 7.5] & $x^*=5.1998$; $f(x^*)=-1.6013$\\
        problem 11 & $2\cos(x)+\cos(2x)$ & [$-\pi/2$, $2\pi$] & $x^*=2.0667$; $f(x^*)=-1.5$ \\
        problem 14 & $-e^{-x}\sin(2\pi x)$ & [0, 4] & $x^*=0.2249$; $f(x^*)=-0.7887$ \\
        problem 15 & $\frac{x^2 - 5x + 6}{ x^2 + 1}$ & [-5, 5] & $x^*=2.4142$; $f(x^*)=-0.0355$ \\
        problem 22 & $e^{-3x} - \sin^3(x)$ & [0, 20] & $x^*=9\pi/2$; $f(x^*)=e^{-27\pi/2}-1$\\
        \hline
    \end{tabular}} 
    \caption{Univariate test problems from \url{https://infinity77.net/global\_optimization}.}
    \label{tab:1}
\end{table}

\newpage

\vspace{-0.25cm}
Table \ref{tab:2} reports the best observed values averaged on the 30 independent runs along with the associated standard deviations. It is evident that every time GP-BO converged to the optimum (i.e., problems 02, 07, 11, and 15) WBGP-BO did the same. On the remaining 5 test problems (i.e., problems 03, 05, 06, 14, and 22) WBGP-BO outperforms GP-BO, almost always significantly ($p$-value<0.05). There is not any relevant difference between WBGP-BO with $N=16$ and $N=32$, but on problem 06. Finally, both the WBGP-BO methods converge to the optimum in problem 14, contrary to GP-BO.
\vspace{-0.25cm}
\begin{table}[h!]
    \centering
    \resizebox{\textwidth}{!}{
    \begin{tabular}{|l||c|cc|cc|}
        \hline
        \;\textbf{test} & \textbf{GP-BO} & \;\textbf{WBGP-BO$_{N=16}$} & $p$-\textbf{value}\; & \;\textbf{WBGP-BO$_{N=32}$} & $p$-\textbf{value}\; \\
        \hline
        \;problem 02\; & \;-1.8996 (0.0000)\; & -1.8996 (0.0000) & - & -1.8996 (0.0000) & - \\
        \;problem 03\; & -8.0563 (3.8520) & \textbf{-10.2932} (1.6467) & 0.007 & -10.2720 (2.2938) & 0.010 \\
        \;problem 05\; & -1.2485 (0.2981) & \textbf{-1.4778} (0.0619) & <0.001 & \textbf{-1.4778} (0.0619) & <0.001 \\
        \;problem 06\; & -0.5667 (0.3387) & -0.6589 (0.3217) & 0.057 & \textbf{-0.7246} (0.2520) & 0.012 \\
        \;problem 07\; & -1.5904 (0.0596) & -1.5904 (0.0596) & - & -1.5904 (0.0596) & - \\
        \;problem 11\; & -1.5000 (0.0000) & -1.5000 (0.0001) & <0.001 & -1.5000 (0.0000) & <0.001 \\
        \;problem 14\; & -0.6128 (0.2494) & \textbf{-0.7887} (0.0000) & <0.001 & \textbf{-0.7887} (0.0000) & <0.001 \\
        \;problem 15\; & -0.0355 (0.0000) & -0.0355 (0.0001) & <0.001 & -0.0355 (0.0001) & <0.001 \\
        \;problem 22\; & -0.9484 (0.1645) & -0.9752 (0.0727) & 0.723 & \textbf{-0.9755} (0.0728) & 0.536 \\
        \hline
    \end{tabular}}
    \caption{Best observed values: average and standard deviation over 30 independent runs. $p$-values refer to a Wilcoxon paired test with GP-BO as baseline. The null hypothesis of the test is that the difference between the paired observations is zero.}
    \label{tab:2}
\end{table}

\newpage 

\begin{figure}[h!]
    \centering
    \includegraphics[width=1.0\linewidth]{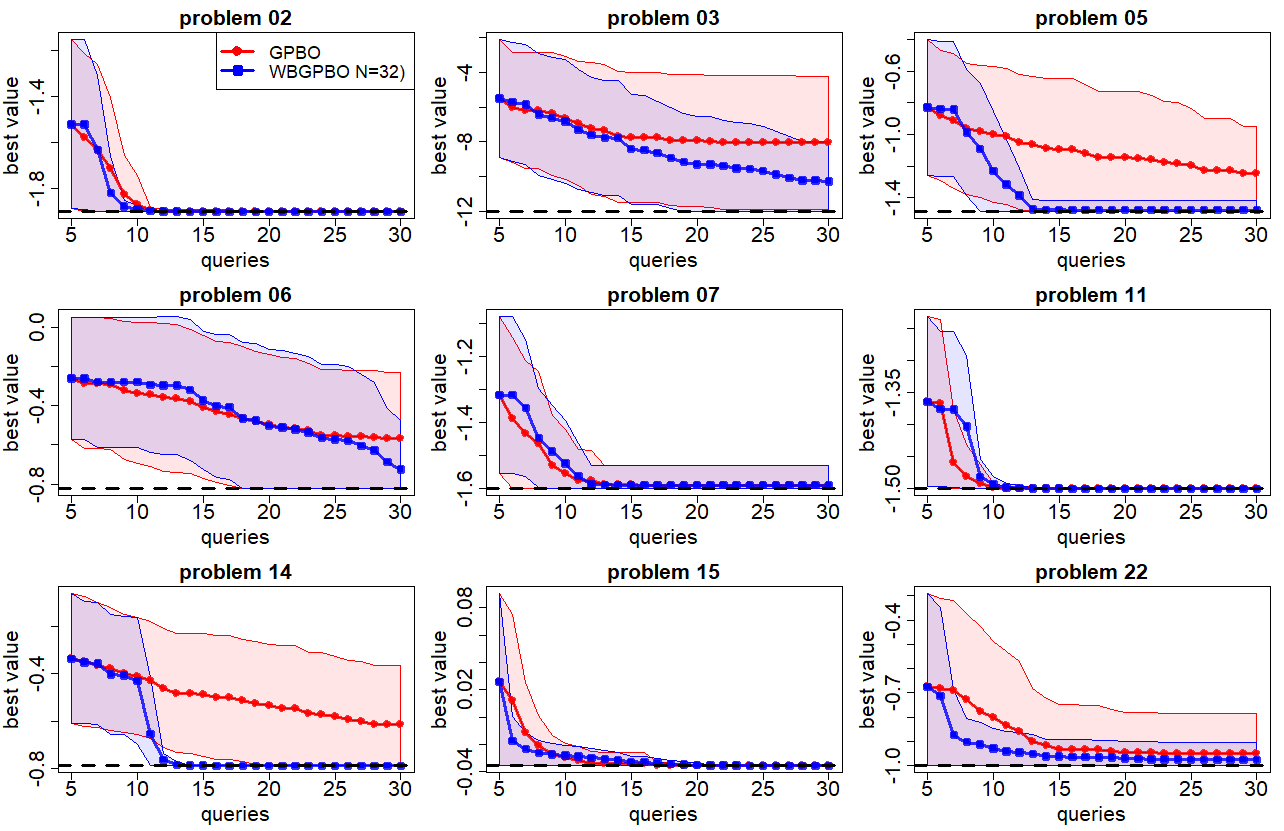}
    \caption{Best observed values over the sequential queries (average and standard deviation over the 30 independent runs). WBGPBO with $N=16$ is omitted because almost always overlapping WBGP-BO with $N=16$ or being slightly worse.}
    \label{fig:charts}
\end{figure}

Figure \ref{fig:charts} shows the evolution of the best observed value over the sequential queries performed by GP-BO and WBGP-BO with $N=32$ (i.e., WBGPBO with $N=16$ is omitted because it almost always overlaps WBGP-BO with $N=16$ or is slightly worse). The ability of WBGP-BO to converge to better solutions than GP-BO, on "difficult" problems is clear, specifically problems 03, 05, 06, 14, and 22, which are characterized by local minima with values close to the global one or resemble the \textit{needle-in-the-haystack} problem.

\section{Conclusions}
In this paper we have presented a novel BO algorithm which uses a set of $N$ predefined kernel's hyperparameters to fit, at every iteration, as many GPs and then combine them into a single GP model obtained as Wasserstein Barycenter of the $N$ GPs. Although preliminary, experiments on onedimensional test problems show promising results, with the proposed WBGP-BO algorithm significantly outperforming GP-BO on the most "tricky" test problems, where vanilla GP-BO get stuck into local minima.

Moreover, we have also explained how the Wasserstein Barycenter GP can be thought  as a particular ensemble of GP models with a statistically sound background.

The most relevant limitation -- we are currently targeting -- regards the set of experiments which has to be extended to include multi-dimensional test problems as well as other competing algorithms like SCoreBO and HE-GP-UCB.

Finally, an interesting perspective regards the possibility to identify a strategy for dynamically setting the weights $\lambda_i$ in the computation of the Wasserstein Barycenter of GPs with the aim to further improve the effectiveness and efficiency of WBGP-BO.

%
%
%
\bibliographystyle{splncs04}
\bibliography{wbgpbo.bib}

\end{document}